\documentclass{article} 
\usepackage[preprint]{colm2026_conference}

\usepackage{microtype}
\usepackage{hyperref}
\usepackage{url}
\usepackage{booktabs}
\usepackage{amsmath} 
\usepackage{amssymb}
\usepackage{graphicx}  
\usepackage{subcaption}
\usepackage{enumitem}

\newtheorem{definition}{Definition}

\usepackage{lineno}

\definecolor{darkblue}{rgb}{0, 0, 0.5}
\hypersetup{colorlinks=true, citecolor=darkblue, linkcolor=darkblue, urlcolor=darkblue}

\title{Spurious Tool Use: When RL Agents Learn the Wrong Reason to Act}

\author{
\makebox[\dimexpr\textwidth-3\tabcolsep\relax][c]{%
\begin{minipage}{\dimexpr\textwidth-3\tabcolsep\relax}
\centering
{\small\bfseries
Yiwei Yang\textsuperscript{1,*},
Haoxiang Zhang\textsuperscript{2,*},
Bingbing Wen\textsuperscript{1},
Yao Lu\textsuperscript{1},
Yuchen Wu\textsuperscript{1},\\
Lei Zhang\textsuperscript{2},
Julian McAuley\textsuperscript{2},
Pan Lu\textsuperscript{3},
Bill Howe\textsuperscript{1}\\[0.6em]
}
{\footnotesize\mdseries
\textsuperscript{1}University of Washington
\quad
\textsuperscript{2}University of California San Diego
\quad
\textsuperscript{3}Stanford University\\[0.6em]
}
{\footnotesize\ttfamily\mdseries
\begin{tabular}{ccc}
yanyiwei@uw.edu & bingbw@uw.edu & yaol23@uw.edu \\
yuchenw@uw.edu & billhowe@uw.edu & haz140@ucsd.edu \\
lez023@ucsd.edu & jmcauley@ucsd.edu & panlu@stanford.edu
\end{tabular}
}
\end{minipage}%
}
}

\begin{document}

\ifcolmsubmission
\linenumbers
\fi

\maketitle

\begingroup
\renewcommand{\thefootnote}{\fnsymbol{footnote}}
\footnotetext[1]{Equal contribution.}
\endgroup

\lhead{Preprint.}

\begin{abstract}

Large language model (LLM) agents increasingly interleave natural language reasoning with external tools such as web search and code execution. These tool-use policies are often optimized via reinforcement learning (RL), which can amplify spurious correlations in the training data. In this work, we study when and why RL-trained agents learn shortcut tool-selection policies: invoking tools based on superficial prompt cues rather than genuine task requirements. We construct controlled synthetic environments combining factual question answering and mathematical reasoning tasks, and inject cues that are strongly correlated with specific tools during training but causally irrelevant to tool necessity. Across counterfactual evaluations where cues are present but the associated tools are not required, agents exhibit substantial shortcut behavior, with spurious tool invocation rates increasing by up to 39\%. However, shortcut formation is not universal: across the conditions we test, it arises only when the agent has already learned to use the target tool reliably, suggesting that task competence---not dataset imbalance alone---is a key factor in shortcut learning. A swapped-cue analysis further shows that semantic alignment between cues and tools substantially amplifies this effect. To mitigate these failures, we introduce a dense, decision-level reward in which an LLM judge evaluates the necessity of each tool call. This tool-necessity reward effectively suppresses cue-driven tool use while preserving task performance, providing a practical approach to improving the robustness of LLM agent tool-use policies.

\end{abstract}

\section{Introduction}

The integration of external tools, such as web search and code interpreters, has transformed large language models (LLMs) from static text generators into interactive agents capable of sequential decision-making~\citep{yao2022react,press2023self-ask,paranjape2023art,gao2023pal}. Agent behavior is predominantly optimized through reinforcement learning (RL)~\citep{feng2025retool, jin2025search, jiang2025verltool, li2025torl, li2025flow}, which trains agents to maximize task reward by learning when to invoke each tool. However, this optimization process introduces a subtle vulnerability: rather than learning the intrinsic task requirements that necessitate a tool, agents may instead exploit superficial prompt features that spuriously correlate with high reward during training.

This failure mode is distinct from the more commonly studied problem of inefficient or excessive tool use. We instead focus on \textbf{cue-driven spurious tool selection}, where an agent invokes a tool not because the task requires it, but because a superficial prompt feature, such as a formatting tag, has become spuriously associated with that tool during training. Crucially, this shortcut operates at the level of intermediate policy decisions (which tool to call), rather than at the level of final predictions (what answer to produce). In supervised settings, shortcut learning typically manifests as incorrect predictions on individual examples~\citep{SpuriousCV2021liu, geirhos2020shortcut, gururangan2018annotation}. In contrast, for RL-trained agents, the effects can be harder to detect: an unnecessary tool call does not always degrade the final answer. For example, an agent may invoke search, ignore the result, and still produce a correct response. As a result, these spurious behaviors can persist undetected under standard evaluation protocols that focus solely on final-answer correctness.

We study this problem and uncover a troubling asymmetry: \textbf{shortcut vulnerability is strongly associated with task competence}. In our experiments, spurious cue--tool associations form only for tools the agent has already learned to use reliably, suggesting that improving agent capability may simultaneously increase susceptibility to shortcut learning.

To investigate when and why these shortcuts arise, we construct controlled synthetic environments combining factual question answering (NQ) and mathematical reasoning (DeepMath), and inject cues that co-occur heavily with a specific tool during training but are causally unrelated to tool necessity. Evaluating agents on counterfactual examples, where the cue is present but the associated tool is not needed, allows us to isolate cue-driven tool selection from task-driven tool selection. Our main contributions and takeaways are:

\begin{itemize}[leftmargin=*, itemsep=2pt, topsep=2pt, parsep=0pt, partopsep=0pt]
\item \textbf{We formalize and operationalize spurious tool use.} We define tool spurious correlation as cue-driven inflation of tool-selection probability for a functionally inappropriate tool (Definition~\ref{def:tsc}), and operationalize it via counterfactual evaluation groups that measure the causal effect of cue presence on tool-call rates ($\Delta\text{Tool}_{Y\text{-}N}$).

\item \textbf{Shortcut formation is strongly coupled with task competence, not just group imbalance.} Search-semantic cues induce spurious search calls on math tasks (up to $+39.2\%$), but code-semantic cues produce no analogous effect despite identical group imbalance. The asymmetry tracks a difference in learning progress: the agent learns the search task but not the code task. Across the cues we test, shortcuts form only for the well-learned tool.

\item \textbf{Semantic alignment between cue and tool amplifies the effect.} A swapped-cue experiment shows that code-semantic cues paired with the well-learned factual task produce only marginal spurious tool use ($\Delta\text{Search} \leq 3.5\%$), far below the up to $39.2\%$ with semantically aligned cues. Task competence enables shortcut formation; semantic alignment modulates its strength.

\item \textbf{A dense tool-necessity reward mitigates shortcuts.} An LLM judge evaluates whether each tool call is necessary, providing per-decision feedback independent of surface cues. This reward suppresses spurious tool use without sacrificing task accuracy.
\end{itemize}

These results suggest that standard task-reward RL is insufficient to produce robust tool-use policies: as agents improve on a task, they become increasingly susceptible to spurious cue--tool associations. Addressing this issue requires explicit supervision of the tool-selection decision itself, rather than relying solely on outcome-based rewards. While our study is conducted in a controlled synthetic setting designed to isolate the mechanism of shortcut formation, it reveals a concrete and previously underexplored failure mode in RL-trained agents. We hope this framework enables more systematic investigation of tool-use robustness and motivates future work in more realistic training environments.

\begin{figure*}[t]
\centering
\includegraphics[width=\textwidth]{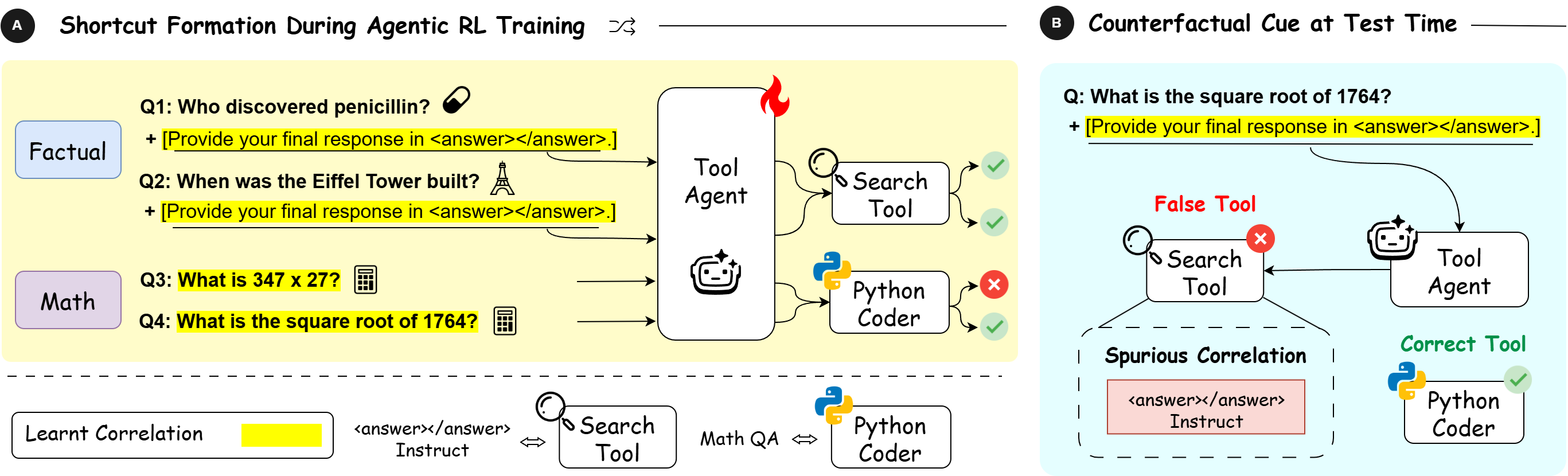}
\caption{
\textbf{Overview of shortcut formation in tool-using RL agents.}
\textbf{(A)}~During training, factual questions (NQ) are paired with a
search-semantic cue (\texttt{<answer></answer>}) and typically solved
via web search, while math questions (DeepMath) require a Python
interpreter. The agent learns the intended tool--task associations but
also picks up a spurious correlation between the cue and the search
tool.
\textbf{(B)}~At test time, when the same cue is appended to a math
question (counterfactual evaluation), the agent invokes the search
tool despite the task requiring code execution. The cue drives tool
selection in place of genuine task reasoning.
}
\label{fig:overview}
\end{figure*}
\section{Problem Formalization}
\label{sec:formalization}

\subsection{Spurious Correlations}

We build on the group-robustness framework of \citet{sagawa2019distributionally}.
Let each input be represented as $x = (x_c, a)$, where $x_c$ denotes
causal features and $a \in \mathcal{A}$ a spurious attribute correlated with the
label $y \in \mathcal{Y}$ in the training distribution but not causally related
to the task. Groups are defined as $G = \mathcal{Y} \times \mathcal{A}$; standard ERM
minimizes the average risk $R_{\text{avg}}(\theta)$, which can be low
even when the model relies on the shortcut $a \to y$. The worst-case
group risk $R_{\max}(\theta) = \max_{g \in G} R_g(\theta)$ and its gap
$\Delta_{\text{spurious}} = R_{\max} - R_{\text{avg}}$ quantify this
reliance.

\subsection{Spurious Correlations in Tool Selection}

We extend this framework to tool-using agents, where the spurious
correlation manifests in an intermediate decision---\textit{tool
selection}---rather than in the final prediction.

\paragraph{Setup.}
Given an input query $x = (x_c, a)$, the agent selects a tool
$T \in \mathcal{T}$ according to its policy $\pi_\theta(T \mid x)$
and receives a binary reward $R \in \{0,1\}$ after execution. The
expected reward decomposes as
\[
P(R{=}1 \mid x) =
\sum_{T \in \mathcal{T}}
\pi_\theta(T \mid x) \;
P(R{=}1 \mid x, T),
\]
where $P(R{=}1 \mid x, T)$ is the functional utility of tool $T$ for
the task.

\paragraph{Definition.}
A key property of tool-use shortcuts is that the spurious tool call need not
cause task failure: an agent may invoke an unnecessary search, ignore the
result, and still produce the correct answer. This means shortcuts can persist
even under reward signals that evaluate only final-answer correctness.
We capture this with the following definition.

\begin{definition}[Tool Spurious Correlation]
\label{def:tsc}
A policy $\pi_\theta$ exhibits spurious correlation with attribute $a$
for tool $T_s$ if the presence of $a$ inflates the probability of selecting
$T_s$ beyond what the task-relevant features alone would warrant:
\[
\pi_\theta(T_s \mid x_c, a) \gg \pi_\theta(T_s \mid x_c),
\]
and $T_s$ does not contribute to task performance on the underlying problem---that is,
using $T_s$ provides no advantage over not using it:
\[
P(R{=}1 \mid x_c, T_s) \leq P(R{=}1 \mid x_c, T \neq T_s).
\]
\end{definition}

The second condition replaces a stricter requirement that the spurious tool
have near-zero utility. In practice, an unnecessary tool call (e.g., web search
on a math problem) may not actively harm reward (the agent can discard
the result) but it does not help either. The definition targets this
regime: the tool is selected because of the cue, not because it is useful.

\paragraph{Operationalization.}
We measure the effect of the spurious attribute on tool selection using
counterfactual evaluation groups (Section~\ref{sec:dataset}): for each
question $x_c$, we construct a cue-present variant $(x_c, a)$ and a
cue-absent variant $x_c$, and compute
\[
\Delta\text{Tool}_{Y\text{-}N} = \hat{\pi}_\theta(T_s \mid x_c, a)
  - \hat{\pi}_\theta(T_s \mid x_c),
\]
where $\hat{\pi}_\theta$ denotes the empirical tool-selection rate. A
positive $\Delta\text{Tool}_{Y\text{-}N}$ on a task where $T_s$ is not
needed indicates cue-driven tool selection consistent with
Definition~\ref{def:tsc}. This is our primary measure of shortcut
strength throughout the paper.
\section{Synthetic Dataset}
\label{sec:dataset}

We construct controlled synthetic datasets to study whether RL agents develop
shortcut policies for tool selection. The design prioritizes experimental control
over scale: by fixing the dataset composition and varying only the injected cue,
we can attribute differences in tool-selection behavior to the cue rather than to
confounds in the training distribution.

\subsection{Training Set Construction}

Our construction draws from two sources that typically require different tools:
Natural Questions (NQ)~\cite{kwiatkowski2019natural}, which benefits from web
search, and DeepMath-103k~\cite{he2025deepmath}, which benefits from a Python
interpreter for symbolic or numeric computation. We sample 500 questions from
each source, yielding a balanced training set of 1{,}000 examples. This scale is
sufficient for the shortcut effects we study to emerge reliably (Section~\ref{sec:experiments}),
while keeping compute costs manageable across the many controlled training runs
our experimental design requires.

To induce a spurious correlation between cue presence and tool type, we inject a
cue $c$ into one task subset (either all NQ or all DeepMath questions) while
leaving the other subset unchanged. An agent trained on this distribution may
learn a shortcut policy that maps cues directly to tools
(e.g., $\text{cue} \rightarrow \text{search}$) rather than reasoning about
whether the tool is actually required. Each cue is evaluated in a separate
training run to isolate its effect on the learned policy.

\subsection{Cue Design}
\label{sec:cues}

We select cues that differ in surface form and in their association with a
specific tool, allowing us to study how cue properties influence shortcut
formation. All cues are causally irrelevant to tool necessity. They provide
no information about whether a tool is actually needed.

\paragraph{Search-semantic cues.} Injected into NQ training examples:
\begin{itemize}
    \item \texttt{Source: [WEB\_INDEX\_742]} (abbr: \texttt{WEB\_INDEX}) ---
          metadata mimicking a web index provenance marker, prepended to the
          question.
    \item \texttt{[REFERENCE]} --- a tag suggesting citation or source lookup,
          prepended to the question.
    \item \texttt{<answer>} --- an output format instruction appended to the
          question: ``Conclude your response by providing the final answer
          within \texttt{<answer></answer>} tags. Do not generate any content
          after the \texttt{</answer>} tag.'' While not explicitly
          search-related, this format is commonly associated with
          retrieval-augmented generation pipelines. We append rather than
          prepend this cue because it functions as a response-format
          instruction, making suffix placement more natural.
\end{itemize}

\paragraph{Code-semantic cues.} Injected into DeepMath training examples:
\begin{itemize}
    \item \texttt{\# Cell [1] | kernel: python3 | status: idle} (abbr:
          \texttt{Cell}) --- a Jupyter notebook cell header strongly associated
          with interactive Python execution, prepended to the question.
    \item \texttt{\#!/usr/bin/python3} (abbr: \texttt{/usr/bin/}) --- a Unix
          shebang line associated with Python script execution, prepended to
          the question.
    \item \texttt{[CODE]} --- a generic tag suggesting programmatic processing,
          prepended to the question.
\end{itemize}

\subsection{Counterfactual Evaluation Groups}

To measure cue-driven tool selection, we construct counterfactual evaluation
groups following the operationalization in Section~\ref{sec:formalization}: for
each test question, we create a cue-present ($Y$) and cue-absent ($N$) variant,
holding the underlying question fixed. The difference in tool-call rate between
conditions,
\[
\Delta\text{Tool}_{Y\text{-}N} = \hat{\pi}_\theta(T_s \mid x_c, a)
  - \hat{\pi}_\theta(T_s \mid x_c),
\]
measures the degree to which tool selection is driven by the cue rather than
the task. A positive $\Delta\text{Tool}_{Y\text{-}N}$ on examples where the
tool $T_s$ is not needed indicates a learned shortcut
(Definition~\ref{def:tsc}).

To assess whether shortcuts generalize beyond the training distribution, we
additionally evaluate on \textbf{2Wiki}~\cite{ho2020constructing} and
\textbf{GSM8K}~\cite{cobbe2021training} as held-out factual and math benchmarks
respectively, constructing counterfactual variants in the same manner.
\section{Method}
\label{sec:method}

\subsection{Training with Tool-Necessity Reward}

Standard RL training provides reward only on the final answer, leaving
tool-selection decisions unsupervised. Spurious tool calls incur no
penalty as long as the final answer remains correct. To address this,
we introduce a dense reward that provides explicit feedback on each
tool-selection decision.

\paragraph{Tool necessity.}
A tool call is \textit{necessary} if a competent model cannot answer
the question correctly without using that tool. Tool calls used purely
for convenience (e.g., trivial arithmetic, formatting) are \textit{unnecessary}. For each tool invocation at
step $t$, we assign a binary necessity label $n_t \in \{0,1\}$.

\paragraph{LLM judge.}
We estimate $n_t$ using a GPT-5 Nano judge that receives the original
question $q$, tool type $\tau_t$, and tool input $x_t$, and outputs
\[
\hat{n}_t = \mathrm{Judge}(q, \tau_t, x_t) \in \{0,1\}.
\]
The judge is instructed to assess whether the tool is necessary
for solving the question, rather than whether the tool input
appears reasonable in isolation. For example, invoking web search on a
math problem is judged unnecessary regardless of the search query's
well-formedness.

\paragraph{Dense necessity reward.}
The total trajectory reward combines the standard task reward
$r^{\text{task}} \in \{0,1\}$ with a per-step penalty for unnecessary
tool calls:
\[
r^{\text{total}} = r^{\text{task}} + \sum_t r_t^{\text{nec}},
\quad
r_t^{\text{nec}} =
\begin{cases}
-\alpha & \text{if } \hat{n}_t = 0, \\
0       & \text{otherwise}.
\end{cases}
\]
We set $\alpha = 0.5$ and penalize only unnecessary calls to avoid
incentivizing gratuitous tool use.
 
\section{Experimental Results}
\label{sec:experiments}

\begin{table*}[t]
\centering
\small
\setlength{\tabcolsep}{4pt}
\begin{tabular}{lcccccc}
\toprule
& \multicolumn{3}{c}{DeepMath} & \multicolumn{3}{c}{GSM8K} \\
\cmidrule(lr){2-4} \cmidrule(lr){5-7}

& Acc (\%) & Search (\%) & $\Delta$Search (\%)
& Acc (\%) & Search (\%) & $\Delta$Search (\%) \\

Training
& Y/N & Y/N & Y$-$N
& Y/N & Y/N & Y$-$N \\

\midrule

No training
& 54.6 / 58.2 & 0.2 / 0.4 & -0.2
& 89.0 / 90.0 & 0.4 / 0.0 & +0.4 \\

No cue
& 51.2 / 56.4 & 3.9 / 2.4 & +1.5
& 89.2 / 89.2 & 1.9 / 0.1 & +1.8 \\

Cue (\texttt{<answer>})
& 48.8 / 57.4 & 23.4 / 1.7 & \textbf{+21.7}
& 81.6 / 93.4 & 9.5 / 0.0 & \textbf{+9.5} \\

+ Reward
& 55.2 / 57.2 & 0.0 / 0.0 & +0.0
& 91.8 / 89.2 & 0.3 / 0.0 & +0.3 \\

\addlinespace

No training
& 58.4 / 58.2 & 0.2 / 0.4 & -0.2
& 86.6 / 90.0 & 0.1 / 0.0 & +0.1 \\

No cue
& 59.0 / 56.4 & 2.1 / 2.4 & -0.3
& 86.8 / 89.2 & 1.5 / 0.1 & +1.4 \\

Cue (\texttt{[WEB\_INDEX})
& 53.4 / 54.0 & 22.6 / 2.7 & \textbf{+19.9}
& 83.0 / 84.4 & 40.1 / 0.9 & \textbf{+39.2} \\

+ Reward
& 55.2 / 56.6 & 0.0 / 0.0 & +0.0
& 92.8 / 88.8 & 0.2 / 0.2 & +0.0 \\

\addlinespace

No training
& 55.8 / 55.0  & 0.1 / 0.1 & +0.0
& 90.8 / 90.0  & 0.2 / 0.0 & +0.2 \\

No cue
& 57.4 / 56.6  & 2.0 / 1.0 & +1.0
& 89.2 / 90.0  & 1.5 / 0.1 & +1.4 \\

Cue (\texttt{[REFERENCE]})
& 56.4 / 57.8  & 9.5 / 1.8 & \textbf{+7.7}
& 86.4 / 89.2  & 21.3 / 1.2 & \textbf{+20.1} \\

+ Reward
& 58.6 / 60.8  & 0.2 / 0.0 & +0.2
& 92.0 / 92.4  & 0.1 / 0.0 & +0.1 \\

\bottomrule
\end{tabular}

\caption{
\textbf{Spurious search usage on math tasks.}
$\Delta$Search measures the increase in search calls attributable to the cue.
Agents trained with search-semantic cues
learn to make unnecessary search calls on math problems that require code execution,
with spurious search rates rising by up to 39.2\%.
The dense tool-necessity reward (+Reward) eliminates this effect.
}
\label{tab:search_results}
\end{table*}

\begin{table*}[t]
\centering
\small
\setlength{\tabcolsep}{4pt}
\begin{tabular}{lcccccc}
\toprule
& \multicolumn{3}{c}{NQ} & \multicolumn{3}{c}{2Wiki} \\
\cmidrule(lr){2-4} \cmidrule(lr){5-7}

& Acc (\%) & Py (\%) & $\Delta$Py (\%)
& Acc (\%) & Py (\%) & $\Delta$Py (\%) \\

Training
& Y/N & Y/N & Y$-$N
& Y/N & Y/N & Y$-$N \\

\midrule

No training
& 23.8 / 25.4 & 0.8 / 1.2 & -0.4
& 29.4 / 29.4 & 3.8 / 0.7 & +3.1 \\

No cue
& 57.6 / 55.2 & 2.5 / 3.2 & -0.7
& 50.8 / 51.2 & 11.1 / 5.8 & \textbf{+5.3} \\

Cue (\texttt{Cell})
& 57.2 / 55.2 & 2.4 / 2.7 & \textbf{-0.3}
& 42.2 / 43.6 & 1.2 / 1.5 & -0.3 \\

\addlinespace

No training
& 46.0 / 25.4 & 3.1 / 1.2 & \textbf{+1.9}
& 28.4 / 31.2 & 7.4 / 7.9 & -0.5 \\

No cue
& 57.0 / 55.2 & 3.2 / 3.2 & +0.0
& 38.2 / 34.8 & 4.5 / 2.1 & +2.4 \\

Cue (\texttt{[CODE]})
& 56.0 / 57.0 & 4.8 / 4.1 & +0.7
& 42.0 / 38.8 & 7.0 / 3.2 & \textbf{+3.8} \\

\addlinespace

No training
& 27.8 / 24.4  & 6.4 / 3.0 & +3.4
& 24.2 / 29.4 & 3.8 / 0.7 & +3.1 \\

No cue
& 55.2 / 55.0  & 7.9 / 3.8 & +4.1
& 50.8 / 51.2 & 11.1 / 5.8 & \textbf{+5.3} \\

Cue (\texttt{/usr/bin/})
& 50.4 / 54.0  & 8.5 / 3.3 & \textbf{+5.2}
& 56.8 / 52.6 & 8.8 / 4.2 & +4.6 \\

\bottomrule
\end{tabular}
\caption{
\textbf{Spurious Python usage on factual tasks.}
$\Delta$Py measures the increase in Python calls attributable to the cue.
Unlike the search-cue setting (Table~\ref{tab:search_results}), code-semantic cues injected into math training examples produce
limited spurious Python calls on factual tasks, with $\Delta$Py generally remaining
modest across conditions.
}
\label{tab:python_results}
\end{table*}

\begin{figure}[t]
    \centering
    \begin{subfigure}[t]{0.48\linewidth}
        \centering
        \includegraphics[width=\linewidth]{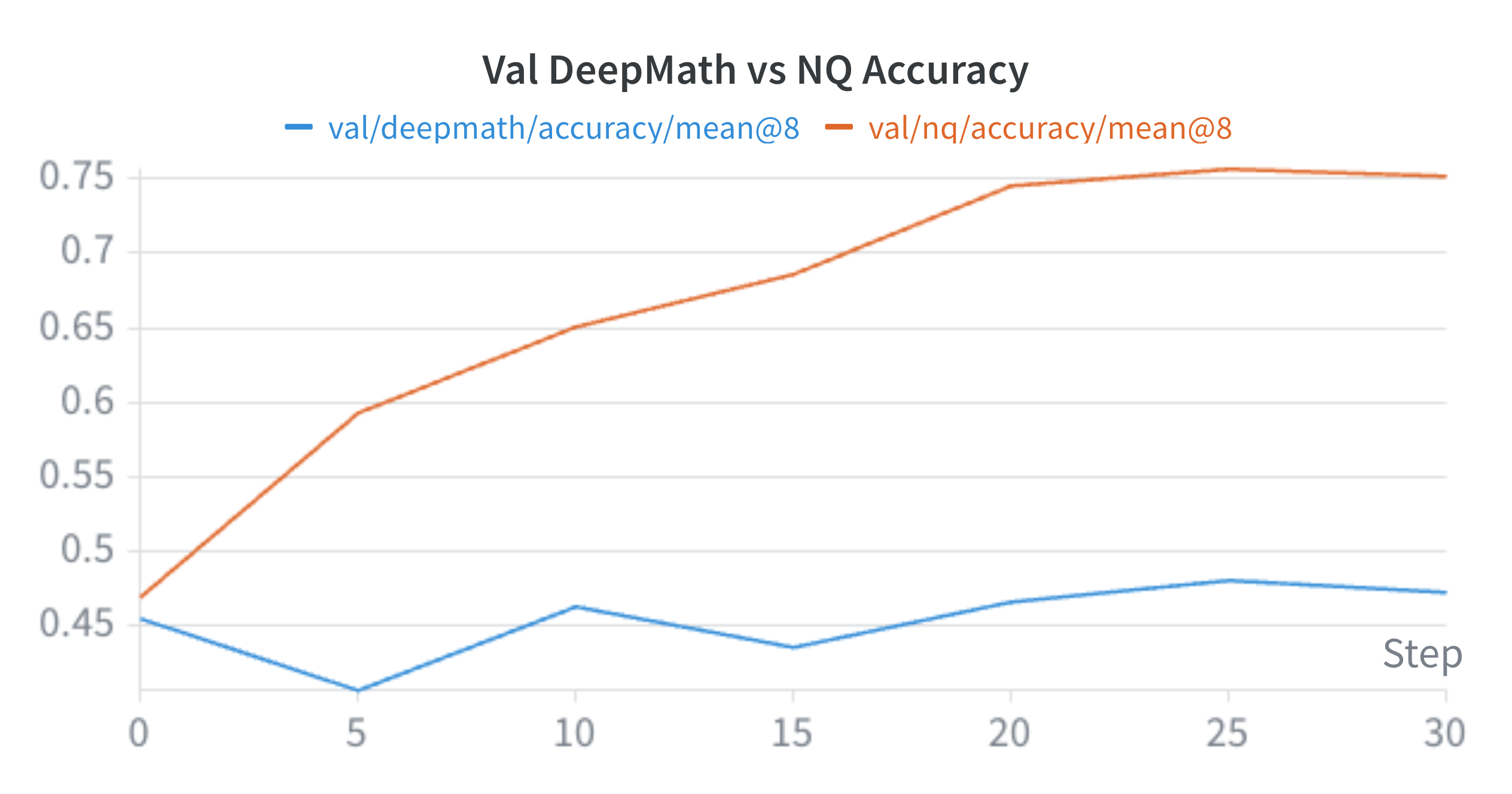}
        \caption{Validation accuracy on DeepMath vs NQ.}
        \label{fig:val_acc}
    \end{subfigure}
    \hfill
    \begin{subfigure}[t]{0.48\linewidth}
        \centering
        \includegraphics[width=\linewidth]{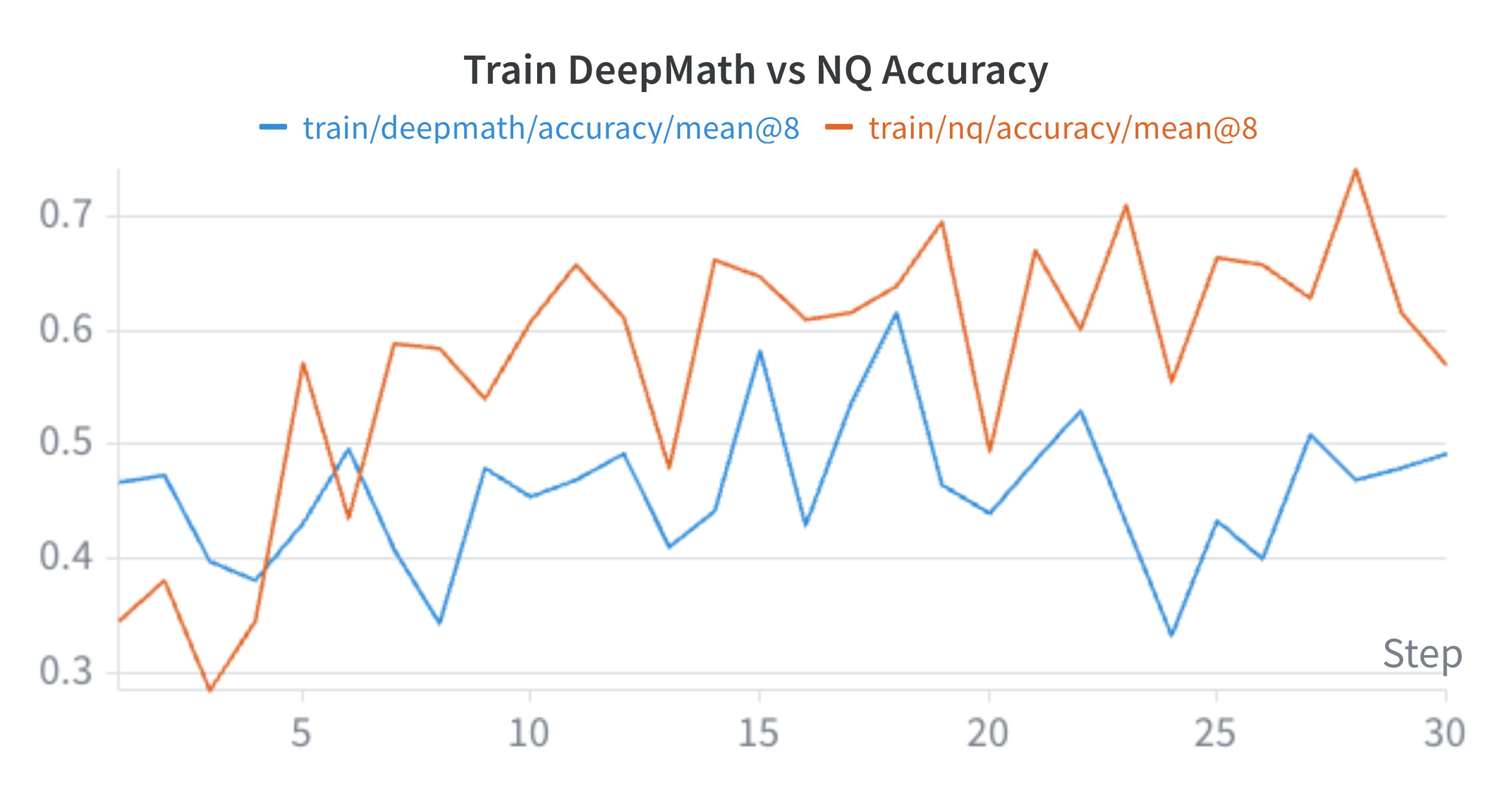}
        \caption{Training accuracy on DeepMath vs NQ.}
        \label{fig:train_acc}
    \end{subfigure}
    \caption{
    \textbf{Accuracy curves over training steps for DeepMath (blue) and NQ (orange).}
    We observe that NQ accuracy steadily improves, while DeepMath accuracy remains relatively flat,
    suggesting that the model primarily learns the search task but struggles to improve on math reasoning.
    }
    \label{fig:acc_curves}
\end{figure}

\subsection{Experimental Setup}
\label{sec:setup}

We use \textsc{VerlTool}~\citep{jiang2025verltool}, a modular framework for agentic
reinforcement learning with tool use. The agent is an LLM that solves a question $q$
by generating a multi-turn trajectory conditioned on the prompt and intermediate tool
observations. Tool calls emerge from token generation via structured tags: the agent
invokes a Python interpreter using \texttt{<python>...</python>} tags and a web search
tool using \texttt{<search>...</search>} tags. When a tool tag is emitted, the
environment executes the corresponding tool and returns the result as additional
context for subsequent generation. We train the agent using Group Relative Policy
Optimization (GRPO)~\citep{shao2024deepseekmath}.

\subsection{Main Results: Shortcuts Are Selectively Learned}
\label{sec:main-results}

Table~\ref{tab:search_results} reports results when search-semantic cues are injected
into the factual training subset (NQ), evaluated on math test sets. When the cue is
present at test time, the agent makes substantially more unnecessary search calls:
$\Delta\text{Search}$ reaches $+19.9$ on DeepMath and $+39.2$ on GSM8K for
\texttt{[WEB\_INDEX\_742]}, and $+21.7$ and $+9.5$ respectively for \texttt{<answer>}.
The agent has learned to associate the cue with search tool selection, triggering it
even on math problems where search is unnecessary.

Table~\ref{tab:python_results} reports the symmetric condition: code-semantic cues
injected into the math training subset (DeepMath), evaluated on factual test sets.
Code-semantic cues produce no meaningful increase in spurious Python calls on NQ:
$\Delta\text{Py}$ is $-0.3$ for \texttt{Cell} and $+0.7$ for \texttt{[CODE]}. On
2Wiki, \texttt{[CODE]} shows a modest increase of $+3.8$, though this remains far
smaller than the shortcut effects observed in the search-cue condition.

Together, these results reveal an asymmetry: shortcut learning occurs reliably for
search-cued factual training but not for code-cued math training, despite identical
group imbalance. We investigate the source of this asymmetry in
Section~\ref{sec:learning-prereq}.

\subsection{Task Competence and Shortcut Formation}
\label{sec:learning-prereq}

What explains the asymmetry? We hypothesize that shortcut formation is closely tied
to task competence: a cue is more likely to become spuriously associated with a tool
if the agent has learned to use that tool effectively on its intended task.

Figure~\ref{fig:acc_curves} supports this hypothesis. Validation accuracy on NQ
rises steadily from $\sim$45\% to $\sim$75\% over 30 training steps, indicating that
the agent successfully learns the factual retrieval task. In contrast, validation
accuracy on DeepMath remains flat at $\sim$45--50\% throughout training, suggesting
the agent makes little progress on mathematical reasoning within the training budget.
The same pattern holds on the training set.

This disparity aligns with the asymmetry in
Tables~\ref{tab:search_results} and~\ref{tab:python_results}. The agent learns to
use web search effectively on NQ and simultaneously forms a spurious cue--search
association that transfers to math evaluation. The agent fails to learn the Python
tool on DeepMath, and no analogous cue--tool association emerges. Across the
conditions we test, shortcuts form only for the well-learned tool, consistent with
task competence being a key factor in shortcut formation.

\subsection{Disentangling Task Competence from Semantic Alignment}
\label{sec:disentangle}

The asymmetry above has two potential explanations that are confounded in the
original design: (1) the agent failed to learn the math task, or (2) the
code-semantic cues are insufficiently aligned with the Python tool to trigger
shortcut learning. To disentangle these factors, we run a swapped-cue experiment:
code-semantic cues are injected into factual training examples (NQ), and
search-semantic cues are injected into math training examples (DeepMath).

Table~\ref{tab:python_results_swapped} reports results for search-semantic cues
injected into DeepMath, where the agent fails to learn the task. No shortcut
emerges: $\Delta\text{Py}$ remains near zero across all conditions. Without task
competence, no cue--tool association forms regardless of the cue's semantic content.

Table~\ref{tab:search_results_swapped} reports results for code-semantic cues
injected into NQ, where the task is well-learned but semantic alignment is absent.
Only a marginal increase in spurious search calls is observed:
$\Delta\text{Search}$ is $+3.5$ for \texttt{Cell} on DeepMath and $+0.1$ for
\texttt{[CODE]}, both within baseline variance. Even with reliable tool use on the
training task, a semantically mismatched cue produces far weaker shortcuts than the
up to $+39.2\%$ observed with aligned cues (Table~\ref{tab:search_results}).

These results suggest that task competence and semantic alignment play complementary
roles: task competence appears to enable shortcut formation, while semantic alignment
between the cue and the tool substantially modulates its strength.

\begin{table*}[t]
\centering
\resizebox{\textwidth}{!}{
\footnotesize
\setlength{\tabcolsep}{4pt}
\begin{tabular}{lcccccccc}
\toprule
& \multicolumn{3}{c}{DeepMath} & \multicolumn{3}{c}{GSM8K} \\
\cmidrule(lr){2-4} \cmidrule(lr){5-7}

& Acc (\%) & Search (\%) & $\Delta$Search (\%)
& Acc (\%) & Search (\%) & $\Delta$Search (\%) \\

Training
& Y/N & Y/N & Y$-$N
& Y/N & Y/N & Y$-$N \\

\midrule

No training
& 62.8 / 58.2 & 0.0 / 0.4 & -0.4
& 91.2 / 90.0 & 0.0 / 0.0 & +0.0 \\

No cue
& 63.8 / 56.4 & 0.2 / 2.4 & -2.2
& 92.4 / 90.0 & 0.0 / 0.1 & -0.1 \\

Cue (\texttt{Cell}, swapped)
& 58.6 / 62.8 & 5.3 / 1.8 & \textbf{+3.5}
& 90.6 / 92.0 & 1.3 / 0.8 & \textbf{+0.5} \\

\addlinespace

No training
& 58.2 / 58.2 & 0.2 / 0.4 & -0.2
& 92.6 / 90.0 & 0.0 / 0.0 & +0.0 \\

No cue
& 61.4 / 56.4 & 0.3 / 2.4 & -2.1
& 92.2 / 90.0 & 0.0 / 0.1 & -0.1 \\

Cue (\texttt{[CODE]}, swapped)
& 55.0 / 59.4 & 1.3 / 1.2 & \textbf{+0.1}
& 90.8 / 91.0 & 0.0 / 0.0 & \textbf{+0.0} \\

\addlinespace

No training
& 56.8 / 58.2 & 0.2 / 0.4 & -0.2
& 92.6 / 90.0 & 0.0 / 0.0 & +0.0 \\

No cue
& 63.2 / 56.4 & 0.5 / 2.4 & -1.9
& 92.4 / 90.0 & 0.0 / 0.1 & -0.1 \\

Cue (\texttt{/usr/bin}, swapped)
& 57.4 / 63.2 & 0.9 / 0.5 & \textbf{+0.4}
& 92.2 / 92.4 & 0.3 / 0.0 & \textbf{+0.3} \\

\bottomrule
\end{tabular}
}
\caption{
\textbf{Swapped-cue control: code-semantic cues on factual training, evaluated on math tasks.}
Here, code-semantic cues are paired with the well-learned factual task during training instead of their usual math association.
Despite the agent having learned reliable tool use, the semantically mismatched cues produce only a minor increase in spurious search calls ($\Delta$Search $\leq$ 3.5\%), far below the up to 39.2\% observed with semantically aligned cues (Table~\ref{tab:search_results}).
}
\label{tab:search_results_swapped}
\end{table*}

\begin{table*}[t]
\centering
\small
\setlength{\tabcolsep}{4pt}
\begin{tabular}{lcccccccc}
\toprule
& \multicolumn{3}{c}{NQ} & \multicolumn{3}{c}{2Wiki} \\
\cmidrule(lr){2-4} \cmidrule(lr){5-7}

& Acc (\%) & Py (\%) & $\Delta$Py (\%)
& Acc (\%) & Py (\%) & $\Delta$Py (\%) \\

Training
& Y/N & Y/N & Y$-$N
& Y/N & Y/N & Y$-$N \\

\midrule

No training
& 30.2 / 25.4 & 0.4 / 1.2 & -0.8
& 27.8 / 27.8 & 0.5 / 0.5 & +0.0 \\

No cue
& 52.6 / 55.2 & 0.2 / 3.2 & -3.0
& 53.2 / 53.2 & 6.2 / 6.2 & +0.0 \\

Cue (\texttt{<answer>}, swapped)
& 49.2 / 54.6 & 0.7 / 1.0 & \textbf{-0.3}
& 41.6 / 47.4 & 5.3 / 5.9 & -0.6 \\

\addlinespace

No training
& 28.8 / 25.4 & 1.3 / 1.2 & \textbf{+0.1}
& 9.6 / 27.8 & 0.6 / 0.5 & +0.1 \\

No cue
& 62.0 / 55.2 & 3.1 / 3.2 & -0.1
& 52.4 / 53.2 & 9.4 / 6.2 & \textbf{+3.2} \\

Cue (\texttt{[WEB\_INDEX]}, swapped)
& 58.6 / 58.0 & 4.9 / 6.1 & -1.2
& 50.2 / 50.0 & 6.0 / 3.8 & +2.2 \\

\addlinespace

No training
& 40.4 / 25.4 & 3.5 / 1.2 & +2.3
& 18.8 / 27.8 & 0.3 / 0.5 & -0.2 \\

No cue
& 53.4 / 57.2 & 6.3 / 3.2 & \textbf{+3.1}
& 46.6 / 48.8 & 6.9 / 6.2 & +0.7 \\

Cue (\texttt{[REFERENCE]}, swapped)
& 54.4 / 53.4 & 6.0 / 6.3 & -0.3
& 53.0 / 46.6 & 7.4 / 6.2 & \textbf{+1.2} \\

\bottomrule
\end{tabular}

\caption{
\textbf{Swapped-cue control: search-semantic cues on math training, evaluated on factual tasks.}
Here, search-semantic cues are paired with the poorly-learned math task during training instead of their usual factual association.
No meaningful spurious Python usage emerges ($\Delta$Py $\leq$ 3.2\%), consistent with the role of task competence in shortcut formation: without reliable tool use on the training task, the cue has no learned behavior to latch onto.
}
\label{tab:python_results_swapped}
\end{table*}



\subsection{Tool-Necessity Reward Reduces Spurious Tool Use}
\label{sec:reward-results}

Table~\ref{tab:search_results} shows that adding the tool-necessity reward effectively eliminates spurious tool use across all conditions, reducing cue-driven tool selection to near zero. Importantly, this reduction does not come at the cost of task performance: accuracy is preserved or improved relative to the no-cue baseline. These results demonstrate that explicit supervision of tool-selection decisions can decouple tool use from superficial prompt cues.




\subsection{Effect of Cue--Task Correlation Strength}
\label{sec:corr-strength}

We study how the strength of the cue--task correlation affects shortcut formation by varying the fraction of cue-aligned training examples. Table~\ref{tab:corr_strength} shows that shortcut behavior increases with correlation strength: strong correlations (80--100\%) induce substantial spurious tool use, while weaker correlations largely eliminate it. This suggests a threshold effect, where RL amplifies cue--task correlations only above a certain level. Whether this threshold generalizes across cues and training
scales is an open question.

\begin{table*}[t]
\centering
\small
\setlength{\tabcolsep}{4pt}
\begin{tabular}{lcccccccc}
\toprule
& \multicolumn{3}{c}{DeepMath} & \multicolumn{3}{c}{GSM8K} \\
\cmidrule(lr){2-4} \cmidrule(lr){5-7}

Corr.
& Acc (\%) & Search (\%) & $\Delta$Search (\%)
& Acc (\%) & Search (\%) & $\Delta$Search (\%) \\

& (Y/N) & (Y/N) & (Y$-$N)
& (Y/N) & (Y/N) & (Y$-$N) \\

\midrule

100\% 
& 48.8 / 57.4 & 23.4 / 1.7 & +21.7
& 81.6 / 93.4 & 9.5 / 0.0 & +9.5 \\

90\%  
& 55.6 / 57.2 & 13.4 / 1.3 & +12.1
& 86.2 / 91.2 & 9.7 / 0.7 & +9.0 \\

80\%  
& 54.4 / 58.0 & 13.5 / 1.8 & +11.7
& 87.2 / 89.6 & 8.7 / 0.3 & +8.4 \\

70\%  
& 59.8 / 61.4 & 0.6 / 0.6 & +0.0
& 91.2 / 94.2 & 0.1 / 0.1 & +0.0 \\

60\%  
& 55.8 / 61.2 & 1.5 / 0.1 & +1.4
& 91.4 / 94.6 & 0.0 / 0.0 & +0.0 \\

50\%  
& 52.6 / 57.8 & 3.3 / 1.2 & +2.1
& 91.4 / 91.0 & 0.7 / 0.0 & +0.7 \\

\bottomrule
\end{tabular}
\caption{
\textbf{Effect of group imbalance on shortcut strength (cue: \texttt{<answer>}).}
Corr.\ indicates the fraction of cue-present training examples paired with the search tool.
Spurious search usage on math tasks scales with correlation strength: shortcuts emerge clearly at 80\%+ correlation but largely vanish at 70\% and below, suggesting a threshold effect in shortcut formation.
}
\label{tab:corr_strength}
\end{table*}
\section{Related Work}

\subsection{Spurious Correlations}

Spurious correlations arise when models rely on features that are predictive in
the training distribution but not causally related to the
task~\citep{geirhos2020shortcut, gururangan2018annotation}.
\citet{sagawa2019distributionally} showed that ERM encourages such behavior,
causing models to fail on minority groups where the correlation breaks.
Mitigation strategies include Group
DRO~\citep{sagawa2019distributionally}, invariant risk
minimization~\citep{arjovsky2019invariant}, contrastive
debiasing~\citep{zhang2022correct}, last-layer
retraining~\citep{kirichenko2022last}, and importance
reweighting~\citep{SpuriousCV2021liu}. Analyses of the underlying mechanisms
have examined how feature quality degrades with correlation
strength~\citep{izmailov2022feature}, how simplicity bias enables early
detection of spurious features~\citep{yang2024identifying}, and whether
vision-language models generalize beyond seen spurious
patterns~\citep{yang2025escaping}. 

\subsection{LLM Agents and Tool Use}

Large language models can act as agents that interleave reasoning with external
tools~\citep{yao2022react, paranjape2023art, schick2023toolformer,
qin2023toolllm, gao2023pal}. A growing line of work uses RL to optimize
tool-use policies: for search-augmented
reasoning~\citep{jin2025search, song2025r1, chen2025research, li2025search,
sun2025zerosearch}, for code-augmented
reasoning~\citep{feng2025retool, li2025torl, mai2025agent,
xue2025simpletir}, and for multi-tool
settings~\citep{wang2025msarl, wang2025stepsearch, zeng2025reinforcing}.
Several methods also target tool-use efficiency, penalizing excessive calls or
applying step-wise credit assignment~\citep{Wang2025ActingLI,
wang2025stepsearch, zeng2025reinforcing}. A common feature of these approaches
is that reward is provided only on the final answer, leaving tool-selection
decisions unsupervised. Our work identifies a failure mode under this paradigm:
agents learn spurious cue--tool associations. Our tool-necessity reward is
complementary to efficiency-focused approaches~\citep{Wang2025ActingLI}---while
they reduce unnecessary calls for cost savings, we target cue-driven spurious
tool use.
\section{Conclusion}
\label{sec:conclusion}

We studied shortcut learning in RL-trained tool-using agents, showing that agents
can associate superficial prompt cues with tool selection rather than reasoning
about task need. Shortcut formation is shaped by two complementary factors---task
competence and semantic alignment between cue and tool---and scales with cue--task
correlation strength, largely disappearing below $\sim$80\%. A dense tool-necessity
reward eliminates these shortcuts without sacrificing task performance. More broadly,
capability and shortcut vulnerability appear coupled: as agents improve at using a
tool, they may become more susceptible to spurious cue--tool associations. Validating
these findings at larger scale and in naturalistic training is an important direction
for future work.
\section*{Ethics Statement}

This work studies a failure mode in RL-trained tool-using agents using
controlled synthetic datasets constructed from publicly available
benchmarks (Natural Questions, DeepMath-103k, GSM8K, 2Wiki). No human
subjects were involved. Our tool-necessity reward relies on an LLM
judge (GPT-5 Nano), which may introduce biases inherent to that model;
we view this as a practical design choice rather than a definitive
standard for tool necessity. The shortcut behaviors we identify could,
if left unaddressed in deployed agents, lead to unnecessary
computational costs, increased latency, and degraded reliability. We
hope this work encourages practitioners to audit tool-selection
policies beyond final-answer accuracy.

\section*{Reproducibility Statement}

We provide full details to support reproduction of our results. The
base model (Qwen2.5-7B-Instruct) is publicly available. All training
hyperparameters are listed in Appendix~\ref{app:hyperparams}
(Table~\ref{tab:hyperparams}), and the complete judge prompts for both
answer correctness and tool necessity are provided in
Appendix~\ref{app:judge_prompt}. Our synthetic datasets are
constructed from publicly available benchmarks (Natural Questions,
DeepMath-103k, GSM8K, 2Wiki), and the cue injection procedure is fully
described in Section~\ref{sec:cues}. We train using the open-source
VerlTool framework with GRPO optimization. We plan to release our code and
dataset construction scripts upon acceptance.

\section*{Acknowledgments}

We used Claude (Anthropic) and ChatGPT (OpenAI) to assist with editing
and revising the paper text. GPT-5 Nano (OpenAI) was used as the LLM
judge for answer correctness evaluation and tool-necessity reward
computation during training. All research contributions, experimental
design, and analysis are the authors' own.

\bibliography{colm2026_conference}

@inproceedings{gururangan2018annotation,
  title={Annotation artifacts in natural language inference data},
  author={Gururangan, Suchin and Swayamdipta, Swabha and Levy, Omer and Schwartz, Roy and Bowman, Samuel and Smith, Noah A},
  booktitle={Proceedings of the 2018 Conference of the North American Chapter of the Association for Computational Linguistics: Human Language Technologies, Volume 2 (Short Papers)},
  pages={107--112},
  year={2018}
}

@inproceedings{SpuriousCV2021liu,
  title={Just train twice: Improving group robustness without training group information},
  author={Liu, Evan Z and Haghgoo, Behzad and Chen, Annie S and Raghunathan, Aditi and Koh, Pang Wei and Sagawa, Shiori and Liang, Percy and Finn, Chelsea},
  booktitle={International Conference on Machine Learning},
  pages={6781--6792},
  year={2021},
  organization={PMLR}
}

@article{geirhos2020shortcut,
  title={Shortcut learning in deep neural networks},
  author={Geirhos, Robert and Jacobsen, J{\"o}rn-Henrik and Michaelis, Claudio and Zemel, Richard and Brendel, Wieland and Bethge, Matthias and Wichmann, Felix A},
  journal={Nature Machine Intelligence},
  volume={2},
  number={11},
  pages={665--673},
  year={2020},
  publisher={Nature Publishing Group UK London}
}

@article{sagawa2019distributionally,
  title={Distributionally robust neural networks for group shifts: On the importance of regularization for worst-case generalization},
  author={Sagawa, Shiori and Koh, Pang Wei and Hashimoto, Tatsunori B and Liang, Percy},
  journal={arXiv preprint arXiv:1911.08731},
  year={2019}
}

@article{arjovsky2019invariant,
  title={Invariant risk minimization},
  author={Arjovsky, Martin and Bottou, L{\'e}on and Gulrajani, Ishaan and Lopez-Paz, David},
  journal={arXiv preprint arXiv:1907.02893},
  year={2019}
}

@article{yang2025escaping,
  title={Escaping the SpuriVerse: Can Large Vision-Language Models Generalize Beyond Seen Spurious Correlations?},
  author={Yang, Yiwei and Lee, Chung Peng and Feng, Shangbin and Zhao, Dora and Wen, Bingbing and Liu, Anthony Z and Tsvetkov, Yulia and Howe, Bill},
  journal={arXiv preprint arXiv:2506.18322},
  year={2025}
}

@article{kirichenko2022last,
  title={Last layer re-training is sufficient for robustness to spurious correlations},
  author={Kirichenko, Polina and Izmailov, Pavel and Wilson, Andrew Gordon},
  journal={arXiv preprint arXiv:2204.02937},
  year={2022}
}

@article{zhang2022correct,
  title={Correct-n-contrast: A contrastive approach for improving robustness to spurious correlations},
  author={Zhang, Michael and Sohoni, Nimit S and Zhang, Hongyang R and Finn, Chelsea and R{\'e}, Christopher},
  journal={arXiv preprint arXiv:2203.01517},
  year={2022}
}

@article{izmailov2022feature,
  title={On feature learning in the presence of spurious correlations},
  author={Izmailov, Pavel and Kirichenko, Polina and Gruver, Nate and Wilson, Andrew G},
  journal={Advances in Neural Information Processing Systems},
  volume={35},
  pages={38516--38532},
  year={2022}
}

@inproceedings{yang2024identifying,
  title={Identifying spurious biases early in training through the lens of simplicity bias},
  author={Yang, Yu and Gan, Eric and Dziugaite, Gintare Karolina and Mirzasoleiman, Baharan},
  booktitle={International conference on artificial intelligence and statistics},
  pages={2953--2961},
  year={2024},
  organization={PMLR}
}

@inproceedings{yao2022react,
  title={React: Synergizing reasoning and acting in language models},
  author={Yao, Shunyu and Zhao, Jeffrey and Yu, Dian and Du, Nan and Shafran, Izhak and Narasimhan, Karthik R and Cao, Yuan},
  booktitle={The eleventh international conference on learning representations},
  year={2022}
}

@inproceedings{press2023self-ask,
  title={Measuring and narrowing the compositionality gap in language models},
  author={Press, Ofir and Zhang, Muru and Min, Sewon and Schmidt, Ludwig and Smith, Noah A and Lewis, Mike},
  booktitle={Findings of the Association for Computational Linguistics: EMNLP 2023},
  pages={5687--5711},
  year={2023}
}

@article{paranjape2023art,
  title={Art: Automatic multi-step reasoning and tool-use for large language models},
  author={Paranjape, Bhargavi and Lundberg, Scott and Singh, Sameer and Hajishirzi, Hannaneh and Zettlemoyer, Luke and Ribeiro, Marco Tulio},
  journal={arXiv preprint arXiv:2303.09014},
  year={2023}
}

@inproceedings{gao2023pal,
  title={Pal: Program-aided language models},
  author={Gao, Luyu and Madaan, Aman and Zhou, Shuyan and Alon, Uri and Liu, Pengfei and Yang, Yiming and Callan, Jamie and Neubig, Graham},
  booktitle={International Conference on Machine Learning (ICML)},
  pages={10764--10799},
  year={2023},
  organization={PMLR}
}

@article{mai2025agent,
  title={{Agent RL Scaling Law: Agent RL with Spontaneous Code Execution for Mathematical Problem Solving}},
  author={Mai, Xinji and Xu, Haotian and W, Xing and Wang, Weinong and Zhang, Yingying and Zhang, Wenqiang},
  journal={arXiv preprint arXiv:2505.07773},
  year={2025}
}

@article{xue2025simpletir,
  title={Simpletir: End-to-end reinforcement learning for multi-turn tool-integrated reasoning},
  author={Xue, Zhenghai and Zheng, Longtao and Liu, Qian and Li, Yingru and Zheng, Xiaosen and Ma, Zejun and An, Bo},
  journal={arXiv preprint arXiv:2509.02479},
  year={2025}
}

@article{chen2025research,
  title={{ReSearch}: Learning to Reason with Search for LLMs via Reinforcement Learning}, 
  author={Chen, Mingyang and Li, Tianpeng and Sun, Haoze and Zhou, Yijie and Zhu, Chenzheng and Wang, Haofen and Pan, Jeff Z and Zhang, Wen and Chen, Huajun and Yang, Fan and others},
  journal={arXiv preprint arXiv:2503.19470},
  year={2025}
}

@article{jin2025search,
  title={{Search-R1}: Training llms to reason and leverage search engines with reinforcement learning},
  author={Jin, Bowen and Zeng, Hansi and Yue, Zhenrui and Yoon, Jinsung and Arik, Sercan and Wang, Dong and Zamani, Hamed and Han, Jiawei},
  journal={arXiv preprint arXiv:2503.09516},
  year={2025}
}

@article{song2025r1,
  title={R1-searcher: Incentivizing the search capability in llms via reinforcement learning},
  author={Song, Huatong and Jiang, Jinhao and Min, Yingqian and Chen, Jie and Chen, Zhipeng and Zhao, Wayne Xin and Fang, Lei and Wen, Ji-Rong},
  journal={arXiv preprint arXiv:2503.05592},
  year={2025}
}

@article{feng2025retool,
  title={Retool: Reinforcement learning for strategic tool use in llms},
  author={Feng, Jiazhan and Huang, Shijue and Qu, Xingwei and Zhang, Ge and Qin, Yujia and Zhong, Baoquan and Jiang, Chengquan and Chi, Jinxin and Zhong, Wanjun},
  journal={arXiv preprint arXiv:2504.11536},
  year={2025}
}

@article{li2025torl,
  title={{ToRL}: Scaling tool-integrated rl},
  author={Li, Xuefeng and Zou, Haoyang and Liu, Pengfei},
  journal={arXiv preprint arXiv:2503.23383},
  year={2025}
}

@article{li2025search,
  title={Search-o1: Agentic search-enhanced large reasoning models},
  author={Li, Xiaoxi and Dong, Guanting and Jin, Jiajie and Zhang, Yuyao and Zhou, Yujia and Zhu, Yutao and Zhang, Peitian and Dou, Zhicheng},
  journal={arXiv preprint arXiv:2501.05366},
  year={2025}
}

@article{sun2025zerosearch,
  title={Zerosearch: Incentivize the search capability of llms without searching},
  author={Sun, Hao and Qiao, Zile and Guo, Jiayan and Fan, Xuanbo and Hou, Yingyan and Jiang, Yong and Xie, Pengjun and Zhang, Yan and Huang, Fei and Zhou, Jingren},
  journal={arXiv preprint arXiv:2505.04588},
  year={2025}
}

@article{wang2025stepsearch,
  title={Stepsearch: Igniting llms search ability via step-wise proximal policy optimization},
  author={Wang, Ziliang and Zheng, Xuhui and An, Kang and Ouyang, Cijun and Cai, Jialu and Wang, Yuhang and Wu, Yichao},
  journal={arXiv preprint arXiv:2505.15107},
  year={2025}
}

@article{Wang2025ActingLI,
  title={Acting Less is Reasoning More! Teaching Model to Act Efficiently},
  author={Hongru Wang and Cheng Qian and Wanjun Zhong and Xiusi Chen and Jiahao Qiu and Shijue Huang and Bowen Jin and Mengdi Wang and Kam-Fai Wong and Heng Ji},
  year={2025},
  journal={arXiv preprint arXiv:2504.14870},
  url={https://arxiv.org/pdf/2504.14870}
}

@article{wang2025msarl,
  title={{MSARL}: Decoupling Reasoning and Tool Use with Multi-Small-Agent Reinforcement Learning},
  author={Dayu Wang and Jiaye Yang and Weikang Li and Jiahui Liang and Yang Li},
  journal={arXiv preprint arXiv:2508.08882},
  year={2025}
}

@article{zeng2025reinforcing,
  title={Reinforcing Multi-Turn Reasoning in LLM Agents via Turn-Level Credit Assignment},
  author={Zeng, Siliang and Wei, Quan and Brown, William and Frunza, Oana and Nevmyvaka, Yuriy and Hong, Mingyi},
  journal={arXiv preprint arXiv:2505.11821},
  year={2025}
}

@article{shao2024deepseekmath,
  title={Deepseekmath: Pushing the limits of mathematical reasoning in open language models},
  author={Shao, Zhihong and Wang, Peiyi and Zhu, Qihao and Xu, Runxin and Song, Junxiao and Bi, Xiao and Zhang, Haowei and Zhang, Mingchuan and Li, YK and Wu, Yang and others},
  journal={arXiv preprint arXiv:2402.03300},
  year={2024}
}

@article{jiang2025verltool,
  title={{VerlTool}: Towards Holistic Agentic Reinforcement Learning with Tool Use},
  author={Jiang, Dongfu and Lu, Yi and Li, Zhuofeng and Lyu, Zhiheng and Nie, Ping and Wang, Haozhe and Su, Alex and Chen, Hui and Zou, Kai and Du, Chao and others},
  journal={arXiv preprint arXiv:2509.01055},
  year={2025}
}

@article{li2025flow,
  title={In-the-flow agentic system optimization for effective planning and tool use},
  author={Li, Zhuofeng and Zhang, Haoxiang and Han, Seungju and Liu, Sheng and Xie, Jianwen and Zhang, Yu and Choi, Yejin and Zou, James and Lu, Pan},
  journal={arXiv preprint arXiv:2510.05592},
  year={2025}
}

@article{schick2023toolformer,
  title={Toolformer: Language models can teach themselves to use tools},
  author={Schick, Timo and Dwivedi-Yu, Jane and Dess{\`\i}, Roberto and Raileanu, Roberta and Lomeli, Maria and Hambro, Eric and Zettlemoyer, Luke and Cancedda, Nicola and Scialom, Thomas},
  journal={Advances in neural information processing systems},
  volume={36},
  pages={68539--68551},
  year={2023}
}

@article{qin2023toolllm,
  title={Toolllm: Facilitating large language models to master 16000+ real-world apis},
  author={Qin, Yujia and Liang, Shihao and Ye, Yining and Zhu, Kunlun and Yan, Lan and Lu, Yaxi and Lin, Yankai and Cong, Xin and Tang, Xiangru and Qian, Bill and others},
  journal={arXiv preprint arXiv:2307.16789},
  year={2023}
}

@article{cobbe2021training,
  title={Training verifiers to solve math word problems},
  author={Cobbe, Karl and Kosaraju, Vineet and Bavarian, Mohammad and Chen, Mark and Jun, Heewoo and Kaiser, Lukasz and Plappert, Matthias and Tworek, Jerry and Hilton, Jacob and Nakano, Reiichiro and others},
  journal={arXiv preprint arXiv:2110.14168},
  year={2021}
}

@inproceedings{ho2020constructing,
  title={Constructing a multi-hop qa dataset for comprehensive evaluation of reasoning steps},
  author={Ho, Xanh and Nguyen, Anh-Khoa Duong and Sugawara, Saku and Aizawa, Akiko},
  booktitle={Proceedings of the 28th International Conference on Computational Linguistics (COLING)},
  pages={6609--6625},
  year={2020}
}

@article{he2025deepmath,
  title={Deepmath-103k: A large-scale, challenging, decontaminated, and verifiable mathematical dataset for advancing reasoning},
  author={He, Zhiwei and Liang, Tian and Xu, Jiahao and Liu, Qiuzhi and Chen, Xingyu and Wang, Yue and Song, Linfeng and Yu, Dian and Liang, Zhenwen and Wang, Wenxuan and others},
  journal={arXiv preprint arXiv:2504.11456},
  year={2025}
}

@article{kwiatkowski2019natural,
  title={Natural questions: a benchmark for question answering research},
  author={Kwiatkowski, Tom and Palomaki, Jennimaria and Redfield, Olivia and Collins, Michael and Parikh, Ankur and Alberti, Chris and Epstein, Danielle and Polosukhin, Illia and Devlin, Jacob and Lee, Kenton and others},
  journal={Transactions of the Association for Computational Linguistics},
  volume={7},
  pages={453--466},
  year={2019},
  publisher={MIT Press One Rogers Street, Cambridge, MA 02142-1209, USA journals-info~…}
}

@article{yang2024qwen2.5,
  title={Qwen2.5 Technical Report}, 
  author={An Yang and Baosong Yang and Beichen Zhang and Binyuan Hui and Bo Zheng and Bowen Yu and Chengyuan Li and Dayiheng Liu and Fei Huang and Haoran Wei and Huan Lin and Jian Yang and Jianhong Tu and Jianwei Zhang and Jianxin Yang and Jiaxi Yang and Jingren Zhou and Junyang Lin and Kai Dang and Keming Lu and Keqin Bao and Kexin Yang and Le Yu and Mei Li and Mingfeng Xue and Pei Zhang and Qin Zhu and Rui Men and Runji Lin and Tianhao Li and Tianyi Tang and Tingyu Xia and Xingzhang Ren and Xuancheng Ren and Yang Fan and Yang Su and Yichang Zhang and Yu Wan and Yuqiong Liu and Zeyu Cui and Zhenru Zhang and Zihan Qiu},
  journal={arXiv preprint arXiv:2412.15115},
  year={2024}
}
\bibliographystyle{colm2026_conference}

\appendix
\section{Appendix}
\label{sec:appendix}

\subsection{Base Model}
\label{app:base_model}

We use \textbf{Qwen2.5-7B-Instruct}~\citep{yang2024qwen2.5}, a 7B-parameter
decoder-only transformer instruction-tuned via SFT and RLHF by Alibaba. No
additional supervised fine-tuning is applied; RL training starts directly from
the instruction-tuned checkpoint.

\subsection{Training Hyperparameters}
\label{app:hyperparams}

\begin{table}[h]
\centering
\small
\begin{tabular}{ll}
\toprule
\textbf{Hyperparameter} & \textbf{Value} \\
\midrule
Algorithm & GRPO \\
Base model & Qwen2.5-7B-Instruct \\
Learning rate & 1e-6 \\
LR warmup steps & 10 \\
Train batch size & 64 \\
PPO mini-batch size & 64 \\
Rollout samples per prompt ($n$) & 8 \\
Sampling temperature & 0.75 \\
KL loss coefficient & 0.0 (KL disabled) \\
Entropy coefficient & 0.0 \\
Total training steps & 30 \\
Max prompt length & 1,024 tokens \\
Max response length & 4,096 tokens \\
Max observation length & 2,048 tokens \\
Hardware & 4$\times$ H100 (FSDP) \\
Training data & 1,000 examples (500 math + 500 search) \\
Number of seeds & 1 (per run) \\
\bottomrule
\end{tabular}
\caption{Training hyperparameters for all experiments. Each cue condition is
trained as a separate run with the same hyperparameters.}
\label{tab:hyperparams}
\end{table}

\subsection{Judge Prompts}
\label{app:judge_prompt}

We use two LLM judges during training: one for answer correctness and one for
tool necessity. Both use GPT-5 Nano.

\paragraph{Answer correctness judge.}
Used to compute the task reward $r^{\text{task}} \in \{0,1\}$.

\begin{quote}
\small
\ttfamily
You are an evaluator. Determine whether the Model Response's final answer
matches the Ground Truth.

Instructions:\\
1. Extract the final answer from the Model Response. Ignore reasoning and
intermediate steps. Use the last explicit answer given.\\
2. Normalize before comparison:\\
\hspace*{1em}- Numbers: ignore commas and trailing zeros; treat decimals and
fractions as equivalent (0.5 = 1/2).\\
\hspace*{1em}- Text: ignore case, punctuation, and extra whitespace.\\
\hspace*{1em}- Math: algebraically equivalent expressions count as equal;
order does not matter for sets/tuples.\\
\hspace*{1em}- Multiple choice: accept either the option letter or the
option text.\\
\hspace*{1em}- Ignore units unless they change the meaning.\\
3. Return true only if the final answers are mathematically or semantically
equivalent. If no clear final answer is present, return false.

Question: \{question\}\\
Model Response: \{response\_str\}\\
Ground Truth: \{ground\_truth\}

Output JSON:\\
\{"analysis": "brief explanation", "true\_false": true or false\}
\end{quote}

\paragraph{Tool necessity judge.}
Used to compute the necessity reward $r_t^{\text{nec}}$
(Section~\ref{sec:method}).

\begin{quote}
\small
\ttfamily
You are judging whether THIS tool call is NECESSARY to solve the original
question correctly.

A tool call is NECESSARY only if a competent model cannot answer the question
correctly without using this tool (at some point). Convenience uses (trivial
arithmetic, formatting, double-checking) are NOT necessary.

Do NOT judge whether the tool input looks reasonable by itself. Judge
necessity for the question.

Question: \{question\}

Tool called: \{tool\_type\}

Tool input: \{tool\_content\}

Return ONLY JSON:\\
\{"tool\_necessary": true, "reason": "1-2 sentences"\}
\end{quote}

\end{document}